\documentclass{article}
\usepackage{ijcai24}

\usepackage{times}
\usepackage{soul}
\usepackage{url}
\usepackage[hidelinks]{hyperref}
\usepackage[utf8]{inputenc}
\usepackage[small]{caption}
\usepackage{graphicx}
\usepackage{amsmath}
\usepackage{amsthm}
\usepackage{booktabs}
\usepackage{algorithm}
\usepackage{algorithmic}
\usepackage[switch]{lineno}

\usepackage{amssymb}
\usepackage{booktabs}
\usepackage{enumitem}
\usepackage{placeins}

\usepackage{amsmath}

\usepackage{svg}
\usepackage{mathtools}

\usepackage{pifont} 
\usepackage{tcolorbox} 

\title{Enabling Mixed Effects Neural Networks for Diverse, Clustered Data Using Monte Carlo Methods}

\author{
Andrej Tschalzev$^1$\and
Paul Nitschke$^2$\and
Lukas Kirchdorfer$^{1,3}$\and
Stefan Lüdtke$^4$\and 
Christian Bartelt$^1$\And
Heiner Stuckenschmidt$^1$\\
\affiliations
$^1$University of Mannheim, Germany\\
$^2$Harvard University, USA\\
$^3$SAP Signavio, Walldorf, Germany\\
$^4$University of Rostock, Germany\\
\emails
\{andrej.tschalzev, christian.bartelt, heiner.stuckenschmidt\}@uni-mannheim.de,
p.nitschke@fas.harvard.edu
lukas.kirchdorfer@sap.com,
stefan.luedtke@uni-rostock.de
}

\begin{document}

\maketitle

\begin{abstract}

Neural networks often assume independence among input data samples, disregarding correlations arising from inherent clustering patterns in real-world datasets (e.g., due to different sites or repeated measurements). Recently, mixed effects neural networks (MENNs) which separate cluster-specific 'random effects' from cluster-invariant 'fixed effects' have been proposed to improve generalization and interpretability for clustered data. However, existing methods only allow for approximate quantification of cluster effects and are limited to regression and binary targets with only one clustering feature. We present \textbf{MC-GMENN}, a novel approach employing \textbf{M}onte \textbf{C}arlo methods to train \textbf{G}eneralized \textbf{M}ixed \textbf{E}ffects \textbf{N}eural \textbf{N}etworks. We empirically demonstrate that MC-GMENN outperforms existing mixed effects deep learning models in terms of generalization performance, time complexity, and quantification of inter-cluster variance. Additionally, MC-GMENN is applicable to a wide range of datasets, including multi-class classification tasks with multiple high-cardinality categorical features. For these datasets, we show that MC-GMENN outperforms conventional encoding and embedding methods, simultaneously offering a principled methodology for interpreting the effects of clustering patterns.

\end{abstract}

\section{Introduction} \label{sec:intro}

Clustering patterns are evident in data across various domains, such as medicine \cite{cafri2019review}, ecology \cite{harrison2018brief}, or e-commerce \cite{fei2021promoting}. 
For instance, in product return forecasting, transaction samples are naturally grouped by customer, product, brand, or geographic location. These clusters can often number in the thousands, with each cluster containing only a small number of samples.
In Deep Neural Networks (DNNs), clustering information is commonly treated as an additional categorical feature, often integrated through numeric encoding (e.g., one-hot encoding) or embeddings ~\cite{hancock2020survey,borisov2021deep}. 
While these approaches improve predictive performance compared to ignoring cluster information, they may encounter issues of overfitting, over-parameterization, and scalability when dealing with high-cardinality categorical features \cite{simchoni2021using}. Furthermore, the models blend cluster information with other features, making it challenging to interpret the specific effects of cluster membership accurately.
In the statistics community, generalized linear mixed models (GLMMs) are well-established for handling clustered data ~\cite{mcculloch2003generalized,pinheiro2006efficient,agresti2012categorical}.
Recently, there has been growing interest in integrating GLMMs with deep learning ~\cite{xiong2019mixed,simchoni2023integrating,nguyen2023adversarially}.
Mixed Effects Neural Networks (MENNs) are partially Bayesian models that use fixed effects DNNs and incorporate clustering features separately as probabilistic random effects.

Existing MENN approaches have demonstrated improved predictive performance and interpretability over conventional encoding and embedding approaches. The main challenge in training MENNs is that the negative log-likelihood loss for classification has no closed-form expression. 
While Markov Chain Monte Carlo (MCMC) methods are common for traditional GLMMs \cite{mcculloch1997maximum,archila2016markov}, modern Bayesian Neural Networks are more frequently trained using variational inference (VI) due to its time efficiency \cite{blei2017variational,jospin2022hands}. 
Consequently, all existing MENN approaches rely on approximate methods like VI, although MCMC could provide an exact quantification of the inter-cluster variance \cite{jospin2022hands}.
This limits the interpretability and thus invalidates the main reason for using mixed effects instead of conventional approaches.

A previously underappreciated fact is that, unlike fully Bayesian neural networks, MENNs only need to sample the parameters of the random effects, which changes the way scalability considerations need to be made.
Moreover, modern MCMC methods, particularly the No-U-Turn Sampler (NUTS) \cite{hoffman2011nuts} greatly speed up convergence of MCMC algorithms compared to the time when MCMC for GLMMs was introduced \cite{mcculloch1997maximum}.
Based on these insights we propose MC-GMENN, an approach to train generalized mixed effects neural networks by combining state-of-the-art MCMC methods and deep learning in an Expectation Maximization (EM) framework (Section \ref{sec:MC-GMENN}). 
In Section \ref{sec:experiments} we demonstrate that our approach:
\begin{itemize}
    \item outperforms existing mixed effects deep learning approaches in performance, time efficiency, and inter-cluster variance quantification (Section \ref{ssec:comp_me}).
    \item scales well to a variety of datasets with high dimensionalities, no. of clustering features, no. of classes, cardinalities, and inter-cluster variance constellations (Section \ref{ssec:multi}).
    \item outperforms encoding and embedding approaches on 16 classification benchmark datasets with multiple high-cardinality categorical features while providing high interpretability (Section \ref{ssec:highcard}).

\end{itemize}

A major factor preventing wider adoption of MENNs is that existing approaches do not apply to classification with multiple high-cardinality clustering features and classes, as we will discuss in Section \ref{sec:related-work}. This paper addresses this gap and, to our knowledge, represents the first empirical demonstration of mixed effects deep learning performance for classification tasks with multiple classes and clustering features.

\section{Monte Carlo Generalized Mixed Effects Neural Networks (MC-GMENN)} \label{sec:MC-GMENN}

In this section, we introduce our generalized model formulation and the proposed Monte Carlo Expectation Maximization (MCEM) training procedure. For a general introduction to GLMMs we refer to Chapter 3 in \cite{agresti2012categorical} and \cite{mcculloch2003generalized}. Existing MENNs and differences to our contribution will be discussed in Section \ref{sec:related-work}.

\subsection{Generalized Mixed Effects Neural Networks} \label{sec:gmenn-def}
Let $\textbf{X} \in \mathbb{R}^{N \times D}$ be the fixed effects design matrix and $\textbf{Y} \in \{0,1\}^{N \times C}$ be a matrix indicating class membership, where $N$ is the number of samples, $D$ is the number of (fixed effects) features, and $C$ is the number of classes. In addition, let 
$\mathbb{Z}$ be a set of random effects design matrices $\textbf{Z}^{(l)} \in \{0,1\}^{N \times Q_l}$ with information about cluster membership for $L$ categorical features of cardinalities $Q_1, ..., Q_L$.

We formulate our GMENN model as:

\begin{equation} \label{eq:model}
    \small
    \textbf{y}_{i} = \phi(f_{\Omega}(\textbf{x}_i) + \sum_l^L \textbf{z}_i^{(l)}\textbf{B}^{(l)})     
\end{equation}

where $f_{\Omega}$ is a neural network parameterized by $\Omega$ and $\textbf{B}^{(l)} \in \mathbb{R}^{Q_l \times C}$ is a matrix with random effect vectors per class for clustering feature $l$. $\phi$ is an activation function depending on whether the target is continuous, binary or multi-class. 
For simplified notation, let $\mathbb{B}$ be the set of all random effects vectors. 


The model is based on the assumptions of traditional GLMMs \cite{mcculloch1997maximum}: 
\begin{enumerate}
    \item The samples $\mathbf{y}_i$ are conditionally independent given the random effects $\mathbb{B}$ and drawn from a distribution $p_{y|\mathbb{B}}$ in the exponential family suitable to describe the target.  
    \item The random effects $\mathbf{b}^{(11)}, ..., \mathbf{b}^{(LC)}$ are assumed to be independent and distributed according to parametric distributions $p_b^{(11)}, ..., p_b^{(LC)}$. Most commonly, Normal distributions with zero mean are used for each $p_b^{(lc)}$: $\textbf{b}^{(lc)} \sim \mathcal{N}(0, \boldsymbol{\Sigma}^{(lc)})$. 
\end{enumerate}

\begin{figure}[tb]
\centering
  \includegraphics[width=0.98\columnwidth]{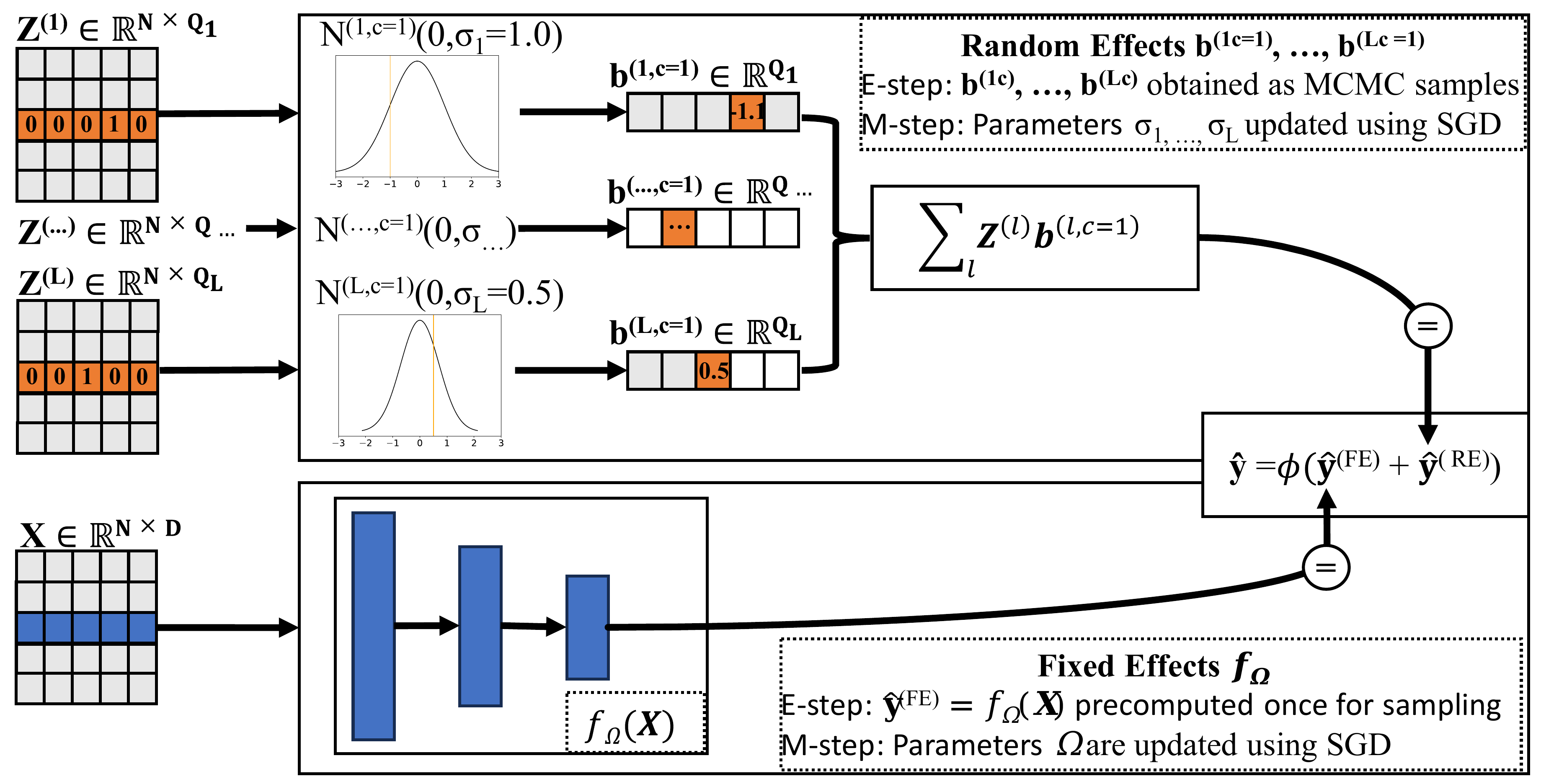}
  \caption[]{Illustration of a generalized mixed effects neural network with MCEM parameter updates for binary classification.}
   \label{fig:system_figure}
\end{figure}

The model is depicted in Figure \ref{fig:system_figure}. By assuming a distribution on the effect of categorical features, the random effects regularize the estimates of cluster effects. Thereby, the amount of regularization (variance parameters) is learned from the data \cite{sigrist2023comparison}.
In the case of the most popular random effects model, the random intercept model, $\boldsymbol{\Sigma}$ simplifies to $\sigma^{2}\textbf{I}$. For simplification, let $\mathbb{S}$ be the set of all covariance matrices.



To fit the model with parameters $\Omega$, $\mathbb{S}$, and $\mathbb{B}$, we need to maximize the marginal data likelihood:


\begin{equation} \label{eq:marginal}
    \small
    \begin{split}
    \mathcal{L}(\Omega, \mathbb{S}|\mathbf{Y}) = \ 
    \prod_{i=1}^{N} \prod_{l=1}^{L} \prod_{c=1}^{C} \int \ 
    p_{y|\mathbb{B}}\left(y_{ic}|\mathbf{b}^{(lc)};\Omega\right) \  \\
    p_b^{(lc)}\left(\mathbf{b}^{(lc)};\boldsymbol{\Sigma}^{(lc)}\right) \ d\mathbf{b}^{(lc)}
    \end{split}
\end{equation}


\subsection{Monte Carlo Expectation Maximization for GMENN} \label{ssec:mcem}
To evaluate the intractable marginal log-likelihood function, we extend the Monte Carlo Expectation Maximization (MCEM) approach for GLMMs \cite{mcculloch1997maximum}\footnote{An implementation of the traditional MCEM algorithm for linear mixed effects regression is available at \nolinkurl{https://www.tensorflow.org/probability/examples/Linear_Mixed_Effects_Models}} to the state-of-the-art of deep learning and MCMC methods. 
For better readability, we use matrix notation throughout the remainder of the paper. In the EM framework, $\textbf{Y}$ represents the observed data, while $\mathbb{B}$ remains unobserved, with unknown variance parameters $\mathbb{S}$.
We substitute Equation \ref{eq:marginal} with a Monte Carlo approximation of its expected value.  

\paragraph{E-Step} 
In each epoch $t$, we create a function to evaluate the expectation of the log-likelihood of Equation \ref{eq:marginal} under the current parameter estimates, given the observed data $\textbf{Y}$: 
\begin{equation} \label{eq:expectation}
    \small
    E^{(t)} = \mathbb{E} \left[ \ln p_{y|\mathbb{B}} \left( \textbf{Y}|\mathbb{B}; \Omega^{(t)} \right) + \ln p_{\mathbb{B}} \left( \mathbb{B}|\mathbb{S}^{(t)} \right) | \textbf{Y}\right]
\end{equation}

where 

\begin{equation} \label{eq:expectation_right}
\small
\ln p_{\mathbb{B}} \left( \mathbb{B}|\mathbb{S}^{(t)} \right) = \sum_{c=1}^C \sum_{l=1}^L \ln p_b^{(lc)} \left( \textbf{b}^{(lc)};{\boldsymbol{\Sigma}^{(lc)}}^{(t)} \right)
\end{equation}

To evaluate this expectation, we would ordinarily need to compute Equation \ref{eq:marginal}, which we aim to avoid. Consequently, we employ Monte Carlo integration to estimate the expectation.
We generate $K$ sets of samples $\mathbb{B}^{(1)}, \ldots, \mathbb{B}^{(K)}$ from the conditional distributions ${p_{b}^{(11)}|\mathbf{Y}, \ldots, p_{b}^{(LC)}|\mathbf{Y} }$ at each epoch $t$. Among these, the initial $R$ samples serve as burn-in. In contrast to existing approaches, we utilize the No-U-Turn Sampler (NUTS) \cite{hoffman2011nuts} for sampling. NUTS is known to demonstrate remarkable efficiency in traversing complex likelihood surfaces, thus significantly reducing the required number of samples \cite{hoffman2011nuts,monnahan2018no}. An ablation study demonstrating the benefits of NUTS compared to other samplers can be found in the supplementary material. Moreover, NUTS can be automated to operate without the need for hyperparameters, making it particularly well-suited for enhancing both the time efficiency and user-friendliness of our MCEM framework.

\paragraph{M-step} In the M-step, we update $\Omega$ and $\mathbb{S}$ using the Monte Carlo estimate of $E^{(t)}$ and gradient descent. Because $\mathbb{B}$ is available as MCMC samples, the two terms in Equation \ref{eq:expectation} can be decoupled:

\begin{subequations} \label{eq:update}
\small
\begin{equation} \label{eq:update_a}
\Omega^{(t+1)} = \Omega^{(t)} + \nabla_\Omega \frac{1}{K-R} \sum_{k=R}^{K} \ln p_{y|\mathbb{B}} \left( \textbf{Y}|\mathbb{B}^{(k)}; \Omega \right)
\end{equation}
\begin{equation} \label{eq:update_b}
\mathbb{S}^{(t+1)} = \mathbb{S}^{(t)} + \nabla_{\mathbb{S}} \frac{1}{K-R} \sum_{k=R}^{K} \ln p_{\mathbb{B}} \left( \mathbb{B}^{(k)};\mathbb{S} \right)
\end{equation}
\end{subequations}


    


To ensure that $f_{\Omega}$ converges and performs well on its own to predict unseen clusters, we add an additional term to Equation \ref{eq:update_a}, inspired by \cite{nguyen2023adversarially}. Hence, the fixed effects loss becomes:  

\begin{equation}
    \small
    \mathcal{L}_{\Omega} = 
    \frac{1}{K-R} \sum_{k}^{K} \ln \ p_{y|\mathbb{B}} \left( \textbf{Y}|\mathbb{B}^{(k)}; \Omega \right) + \lambda \ln p_{y|\mathbb{B}} \left( \textbf{Y}|\mathbf{0}; \Omega \right) 
\end{equation}

where $\mathbf{0}$ denotes the random effects set to zero, and $\lambda$ is a hyperparameter that controls how much emphasis should be placed on the fixed effects.

Convergence of MC-GMENN is determined by early stopping on validation data using Equation \ref{eq:update_a} or a performance metric. 
After convergence, 
estimations of the random effects $\hat{\textbf{b}}^{(lc)}$ are obtained as the mean of all samples over all epochs. 
Predictions are obtained using the estimated random effects coefficients in the model (Equation \ref{eq:model}).

\paragraph{Important MCEM Properties for Deep Learning}
Three key properties make the MCEM procedure exceptionally suitable for combining mixed effects and deep learning: 
First, the sampling (E-step) is detached from the mini-batch loss evaluation (M-step). In the E-step, the computation of $f_{\Omega}(\textbf{X})$ for the first term of Equation \ref{eq:expectation} is required only once. Hence, the E-step can be evaluated efficiently without necessitating mini-batches. In the M-step, the random effects $\mathbb{B}$ are available as MCMC samples. Hence, no expensive integration procedure limits the speed of the mini-batch gradient descent updates.
Consequently, the algorithm scales well to large neural networks, high fixed effects feature dimensionality and sample size. 
Second, the updates in Equations \ref{eq:update_a} and \ref{eq:update_b} can be computed independently. Depending on the specific task, varying update policies and early stopping rules can be used for Equation \ref{eq:update_b}. In the case of the Gaussian random intercept model, which is the most popular choice, the variance parameters $\mathbb{S}$ are updated by setting them to the variance of the current epoch's samples. If exact variance estimation is required, the training can be continued by only updating \ref{eq:update_b} until the variance parameters converge.
Third, Equation \ref{eq:update_a} aligns with the conventional loss function formulation, facilitating the use of common losses that have a corresponding log-likelihood counterpart such as cross-entropy during the M-step. Furthermore, Equation \ref{eq:update_a} naturally decomposes into mini-batches, which is not the case for all MENNs.

\paragraph{Hyperparameters and Automation}
The additional hyperparameters added by MC-GMENN are the (initial) step size $\epsilon$ of NUTS, the no. of samples $K$, the no. of epochs to use as burn-in $R$ and the fixed effects weight $\lambda$.
A too large $\epsilon$ exponentially increases the probability of rejection, while a too small $\epsilon$ is very time-consuming. 
We found that common step size adaption methods like dual averaging \cite{hoffman2011nuts} do not perform well. Instead, we propose to start with a large step size of $\epsilon_0 = 0.1$ and divide it by two whenever the acceptance rate gets below $0.001$. 
Due to the ability of NUTS to traverse large distances, one informative sample per epoch can suffice for a good estimation.
Moreover, a different configuration than $\lambda=1$ is only required if the fixed effects exhibit significantly slower convergence compared to the random effects.
In Section \ref{sec:experiments}, we show that no hyperparameter optimization is necessary to achieve competitive results with our method as it consistently demonstrates strong performance across a diverse range of datasets.


\section{Experiments} \label{sec:experiments}


In all experiments, we compare MC-GMENN to the following conventional methods: Ignoring high-cardinality categorical features (\textit{Ignore}),  one-hot-encoding (\textit{OHE}), target encoding (\textit{TE}) \cite{micci2001preprocessing}, and  entity embeddings (\textit{Embedding}) \cite{guo2016entity}.
For each method, we use the same base neural network architecture and optimizer with the same hyperparameters (learning rate, decay, dropout, embedding size), as well as training procedure (epochs, patience, batch size) per experiment.
To prove that our method is readily applicable to any dataset, we use the default hyperparameters for MC-GMENN over all experiments: $\epsilon_0=0.1, K=2, R=1, \lambda=1$.
In line with previous work \cite{simchoni2023integrating,nguyen2023adversarially}, we use the area under the ROC curve (AUC) as the performance metric. 
To be able to compare across different datasets, training time is evaluated relative to the \textit{Ignore} method.
For simulated datasets, we additionally evaluate the ability of the MENN models to learn the underlying random effects distribution. For that, we use the absolute error of the estimated variance components
and visually compare the learned distributions. 
Our framework and all the evaluated models are implemented in TensorFlow.\footnote{Our code is available at \url{https://github.com/atschalz/mcgmenn}}
More detailed information about the datasets, hyperparameters, and evaluation setup are provided in the supplementary material.




\subsection{Comparison of MC-GMENN with Related Mixed Effects Approaches} \label{ssec:comp_me}

\paragraph{Experimental Setting}
To achieve a fair evaluation of our method and existing approaches, we replicate the binary classification experiments of LMMNN, one of the most recent MENN approaches \cite{simchoni2023integrating}. 
Using the data generation method described in their paper, we generate datasets with features nonlinearly related to the target and additional categorical clustering features with specified cardinalities. 
For the latter, the target varies according to a Normal distribution with specified variance depending on the cluster membership. 
Nine scenarios are simulated for five iterations with varying $Q \in \{100, 1000, 10000\}$ and  $\sigma^2 \in \{0.1, 1.0, 10.0\}$.
The utilized neural network architecture consists of four hidden layers with 100, 50, 25, and 12 neurons with ReLU activation and dropout regularization of 25\%.
Furthermore, our evaluation incorporates target encoding, MC-GMENN, and ARMED \cite{nguyen2023adversarially}, with the latter representing another recent and noteworthy MENN approach.
For ARMED, we use the same hyperparameters as \citeauthor{nguyen2023adversarially} [\citeyear{nguyen2023adversarially}] in their simulated experiments. 

\begin{table}[tb]
    \centering
    \small
    \begin{tabular}{lrrrr}
    \toprule
    {} &   \vtop{\hbox{\strut MC-GMENN}\hbox{\strut (ours)}} &   LMMNN &   ARMED & TE  \\
    \midrule
    MRR   $\uparrow$ &  \textbf{0.63} &  0.60 &  0.16 &  0.55 \\
    Diff. \% $\downarrow$   &  \textbf{0.21} &  0.33 &  8.12 &  0.93 \\
    Train time  $\downarrow$ &  \textbf{0.92} &  7.90 &  4.81 &  1.03 \\
    MAE($\hat\sigma^2$) $\downarrow$ &  \textbf{0.21} &  0.75 &  0.79 &     - \\
    \bottomrule
    \end{tabular}
    \caption{Comparison of mixed effects deep learning approaches on the simulated datasets used in the LMMNN paper \protect\cite{simchoni2023integrating}. The results are averaged over the nine simulated datasets and five iterations. ’MRR’ denotes the mean reciprocal rank and ’Diff \%.’ is the average relative difference in \% of a method compared to the best method in terms of AUC. Train time is reported in minutes relative to the 'Ignore' method. MAE($\hat\sigma^2$) denotes the mean absolute error of the estimated variance components. Only the best-performing non-mixed effects approach is reported (\textit{TE}). Full results can be seen in the supplementary material.}
    \label{tab:me-comp-results}
\end{table}

\paragraph{MC-GMENN Outperforms Existing Mixed Effects Approaches}
As can be seen in Table \ref{tab:me-comp-results}, MC-GMENN shows the best overall performance. 
Compared to the reported results in \citeauthor{simchoni2023integrating} [\citeyear{simchoni2023integrating}], our replication shows very similar performances, training times, and variance quantification (see supplementary material). This indicates that our replication was correct and fair towards LMMNN and strengthens the superiority of our proposed approach.
Unexpectedly, our findings indicate that ARMED underperforms in high-cardinality scenarios which will be further elaborated in Section \ref{sec:related-work}.

\begin{figure}[tb]
  \centering
  \small
  \includegraphics[width=0.9\columnwidth]{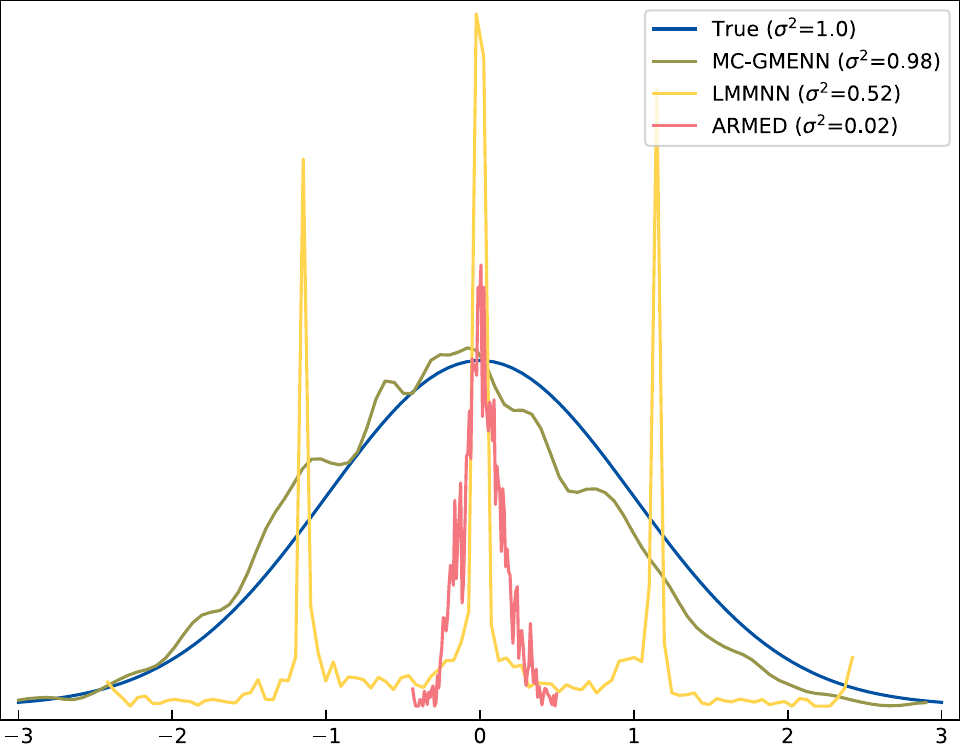}
  \caption{Comparison of the learned random effects distribution of different mixed effects neural network approaches on a simulated dataset with one clustering feature ($Q=1000$ and $\sigma^2=1.0$).}
   \label{fig:re_comparison}
\end{figure}

\paragraph{MC-GMENN Precisely Quantifies Inter-Cluster Variance}
Figure \ref{fig:re_comparison} shows that only our MC-GMENN approach is able to fit the posterior distribution correctly. LMMNN uses a Gaussian quadrature approximation which forces the random effects close to the quadrature points, leading to a bad approximation. 
At the same time, the training time is very high already, such that refining the approximation by using more quadrature points is infeasible, especially in high-cardinality settings.
It can also be seen that ARMED, which uses a VI approach, is strongly biased towards the prior variational distribution ($\sigma^2=0.01$). This hinders the correct estimation of larger cluster effects. 
The result is a worse posterior estimation and ultimately a worse performance. 
In contrast, MC-GMENN provides an unbiased estimate of the integral and thus is able to learn the underlying distribution correctly.

\begin{table*}[tb]
    \centering
    \small
    \begin{tabular}{llllll}
    \toprule
    Variation &         MC-GMENN (ours) &            TE &           OHE &     Embedding &        Ignore \\
    \midrule
    \textbf{Base}:             & \textbf{0.77 (0.011)} &  0.63 (0.013) &  0.74 (0.014) &  0.75 (0.014) &  0.62 (0.007) \\
    \textbf{1M. samples}: $N=1,000,000$       &  \textbf{0.78 (0.015)} &   0.62 (0.01) &  \textbf{0.78 (0.015)} &  \textbf{0.78 (0.014)} &   0.63 (0.01) \\
    \textbf{High-dimensionality}: $D=1,000$    &  \textbf{0.55 (0.012)} &   0.5 (0.003) &  0.52 (0.008) &  0.52 (0.003) &   0.5 (0.003) \\
    \textbf{100 classes}: $C=100$      &  \textbf{0.71 (0.004)} &  0.52 (0.011) &  0.63 (0.006) &  0.64 (0.013) &  0.58 (0.003) \\
    \textbf{High-cardinality}: $Q_1=Q_2=Q_3=20,000$ & \textbf{0.61 (0.011)} &  0.56 (0.013) &     -- &  \textbf{0.59 (0.004)} &  \textbf{0.62 (0.004)} \\
    \textbf{Dominant REs}: $\sigma^2_1=5.0$       &  \textbf{0.91 (0.002)} &  0.58 (0.021) &   0.9 (0.002) &  \textbf{0.91 (0.002)} &  0.58 (0.006) \\
    \textbf{Irrelevant REs}: $\sigma^2_1=\sigma^2_2=\sigma^2_3=0.0001$    &  0.64 (0.007) &  \textbf{0.64 (0.009)} &   0.6 (0.007) &  0.61 (0.006) &  \textbf{0.65 (0.006)} \\
    \textbf{Variance-per-class}: $\sigma^2_2=[.0001,.25,.5,.75,.5]$ &  \textbf{0.75 (0.012)} &  0.63 (0.008) &   0.72 (0.01) &  0.73 (0.011) &  0.63 (0.007) \\
    \textbf{10 REs}: $L=10$ with $Q_1=Q_2=\ldots=Q_L=$\\ $1000$ and  $\sigma^2_1=0.1, \sigma^2_2=0.2, \ldots,  \sigma^2_L=1.0$            &   \textbf{0.9 (0.001)} &  0.55 (0.011) &  0.88 (0.002) &  0.89 (0.002) &  0.58 (0.005) \\
    \midrule
    \midrule
    MRR $\uparrow$ & \textbf{0.87} &  0.26 &  0.34 &      0.44 &   0.40 \\
   Diff. \% $\downarrow$ &  \textbf{0.23} &  19.56 &  4.73 &      3.65 &   17.13 \\
        \bottomrule
    \end{tabular}
    \caption{Performance comparison (AUC) for simulated multi-class classification datasets with multiple clustering features. The results are averaged over five iterations. ’MRR’ denotes the mean reciprocal rank and ’Diff \%.’ is the average relative difference in \% of a method compared to the best method in terms of AUC. Results for the best method and results that do not significantly differ in a paired t-test ($\alpha=0.05$) are highlighted.}
    \label{tab:multi-scenarios}
\end{table*}

\paragraph{MC-GMENN is Time-Efficient Compared to Existing Mixed Effects Approaches}
The training time comparison in Table \ref{tab:me-comp-results} shows that MC-GMENN is more efficient than the competitors. 
For ARMED, high-cardinality features lead to a parameter explosion greatly increasing the training time, as will be discussed in Section \ref{sec:related-work}. 
In LMMNN, the loss function contains matrix inversions that are evaluated in mini-batches making it inefficient.
In contrast, the MCEM procedure of MC-GMENN separates the expensive part in the E-step such that the mini-batch updates in the M-step are as time-efficient as with a regular neural network. 


\subsection{Scalability to Multi-Class Datasets with Multiple Clustering Features} \label{ssec:multi}

\paragraph{Experimental Setting} 
To evaluate the performance of MC-GMENN when applied to datasets with multiple clustering features and classes, we extend the data generation algorithm of \citeauthor{simchoni2023integrating} [\citeyear{simchoni2023integrating}] to multi-class classification. 
As will be discussed in Section \ref{sec:related-work}, LMMNN and ARMED are not applicable to these datasets. 
To demonstrate the scalability and versatility of MC-GMENN, we simulate datasets for one realistic base scenario with five repetitions and vary this scenario according to relevant data properties. 
The base scenario is defined as a multi-class classification task with $N=100,000$, $D=10$, $C=5$, and three clustering features with $Q_1=1000$, $Q_2=10$, $Q_3=1000$. The variance is assumed to be constant per class with $\sigma^2_1=0.0001$, $\sigma^2_2=0.5$, $\sigma^2_3=0.5$. Hence, there is one categorical feature with no impact and two with medium impact on the target, one of which has low and one has high cardinality. 
Further scenarios are simulated with five repetitions each by varying the base scenario as described in Table \ref{tab:multi-scenarios}.
The same base neural network architecture as in Subsection \ref{ssec:comp_me} is used.

\paragraph{MC-GMENN Consistently Matches or Outperforms Encoding and Embedding Approaches} As can be seen in Table \ref{tab:multi-scenarios}, MC-GMENN shows the best average performance across all scenarios. Even in scenarios where it is not the single best model, the relative difference to the best model is low.  
Moreover, Table \ref{tab:multi-scenarios-timevar} shows that MC-GMENN quantifies the inter-cluster variance in multi-class settings with low mean absolute differences to the true variances.


\begin{table}[tb]
    \centering
    \small
    \begin{tabular}{rrrr}
    \toprule
    {} &         Train time $\downarrow$ &            MAE($\hat\sigma^2$) $\downarrow$ \\ 
    \midrule
    Base             &  1.13 &  0.11 \\
    1M. samples        &  2.32 &  0.11 \\
    High-dimensionality     &  1.68 &  0.32 \\
    100 classes       &  4.87 &  0.18 \\
    High-cardinality &  2.46 &  0.55 \\
    Dominant REs        &  2.33 &  0.31 \\
    Irrelevant REs     &  2.67 &  0.01 \\
    Variance-per-class &  1.73 &  0.09 \\
    10 REs             &  2.36 &  0.11 \\
    \bottomrule
    \end{tabular}
    \caption{Comparison of training time and inter-cluster variance quantification of MC-GMENN across different scenarios. Training time for each condition is reported relative to the training time of a neural network where clustering features are ignored. The results are averaged over five iterations.}
    \label{tab:multi-scenarios-timevar}
\end{table}

\paragraph{MC-GMENN Scales to Various Data Scenarios}
The superior performance proves, that MC-GMENN is applicable to different data dimensionality or clustering variance structures. 
As can be expected, the training time relative to the \textit{Ignore} condition increases for more complex scenarios. 
It can be observed that the relative increase in training time is higher for datasets with more complex random effects structures, such as the 100 classes scenario and the high-cardinality scenario. 
The relative training time for 1000 features is almost the same as for 10 features and does not greatly increase with 1 million samples.
This demonstrates that, as discussed in Subsection \ref{ssec:mcem}, MC-GMENN scales well to large datasets. 

\subsection{Application to Diverse Real-World Datasets} \label{ssec:highcard}

\begin{table*}[tb]
    \centering
    \small
    \begin{tabular}{rrr|rrrrr}
    \toprule
    {} &  N/ D/ C & L/ $Q_{max}$  & \vtop{\hbox{\strut MC-GMENN}\hbox{\strut (ours)}} &            TE &           OHE &     Embedding &        Ignore \\
    \midrule
    kdd\_internet\_usage   &  10,108/ 62/ 2 & 5/ 28   &  0.94 (0.003) &  0.94 (0.005) &  \textbf{0.94 (0.007)} &  \textbf{0.95 (0.006)} &  0.94 (0.004) \\
    adult         &  48,842/ 11/ 2 & 3/ 41          & \textbf{ 0.91 (0.003)} &  \textbf{0.91 (0.001)} &   0.9 (0.002) &   0.9 (0.002) &   0.9 (0.001) \\    
    churn      &  5000/ 17/ 2 & 2/ 51           &   0.88 (0.02) &  0.88 (0.018) &   0.9 (0.024) &   \textbf{0.9 (0.024)} &  0.72 (0.046) \\
    porto-seguro    &  595,212/ 54/ 2 & 4/ 104          &  \textbf{0.56 (0.004)} &  \textbf{0.55 (0.004)} &  0.55 (0.005) &  0.54 (0.004) &  \textbf{0.55 (0.005)} \\
    kick            &  72,983/ 23/ 2 & 9/ 1,063          &  0.74 (0.011) &  \textbf{0.75 (0.007)} &  0.71 (0.007) &  0.72 (0.007) & \textbf{0.75 (0.006)} \\ 
    open\_payments      &  73,354/ 1/ 2 & 4/ 4,365       &  \textbf{0.93 (0.009)} &  \textbf{0.92 (0.006)} &  0.91 (0.004) &  \textbf{0.92 (0.007)} &  0.49 (0.007) \\    
    Amazon\_employee &  32,769/ 0/ 2 & 9/ 7,518 &  \textbf{0.84 (0.009)} &  0.81 (0.022) &  \textbf{0.83 (0.005)} &   \textbf{0.84 (0.01)} &     0.5 (0.0) \\
    KDDCup09\_upselling  &  50,000/ 188/ 2 & 20/ 15,415    &   \textbf{0.8 (0.013)} &   0.77 (0.01) &  0.78 (0.013) &   0.7 (0.016) &   \textbf{0.79 (0.01)} \\    
    road-safety-drivers-sex  &  233,964,/ 4/ 2 & 2/ 20,397  &  \textbf{0.73 (0.004)} &  0.72 (0.002) &  \textbf{0.73 (0.003)} &  0.71 (0.002) &   0.7 (0.003) \\
    Click\_prediction\_small &  39,948/ 3/ 2 & 6/ 30,114 &  \textbf{0.66 (0.009)} &  0.63 (0.013) &  0.61 (0.019) &   0.62 (0.02) &  0.62 (0.005) \\

    \midrule
    hpc-job-scheduling  &  4,331/ 6/ 4 & 1/ 14  &  0.91 (0.008) &  0.71 (0.072) &  \textbf{0.92 (0.011)} &  0.85 (0.103) &  0.69 (0.089) \\    
    eucalyptus    &  736/ 15/ 5 & 4/ 27       &   \textbf{0.9 (0.022)} &   \textbf{0.9 (0.023)} &  \textbf{0.89 (0.032)} &  \textbf{0.91 (0.026)} &  \textbf{0.91 (0.031)} \\
    video-game-sales  &  16,598/ 6/ 12 & 2/ 578   &  \textbf{0.79 (0.009)} &    0.6 (0.02) &  0.77 (0.006) &  0.78 (0.006) &   0.7 (0.011) \\    
    Diabetes130US   &  101,766/ 40/ 3 & 7/ 790     &  \textbf{0.65 (0.002)} &  0.54 (0.035) &  0.61 (0.004) &  0.61 (0.002) &  0.62 (0.005) \\    
    Midwest\_survey   &  2,778/ 25/ 10 & 1/ 1008    &  \textbf{0.88 (0.023)} &   0.74 (0.01) &  0.82 (0.006) &  \textbf{0.88 (0.013)} &   0.75 (0.01) \\
    okcupid-stem    &  50,789/ 8/ 3 & 11/ 7019     &  \textbf{0.8 (0.004)} &  0.62 (0.019) &  0.75 (0.004) &   0.74 (0.01) &  0.73 (0.005) \\

    \midrule
    \midrule
    \multicolumn{3}{r|}{Mean reciprocal rank $\uparrow$} &   \textbf{0.72} &  0.40 &  0.37 &     0.45 &  0.34 \\
    \multicolumn{3}{r|}{Mean diff. to best model in \% $\downarrow$} &  \textbf{0.47} &  7.43 &  2.74 &     3.28 &  11.53 \\
    \bottomrule
    \end{tabular}
    \caption{Performance comparison (AUC) for diverse real-world classification datasets. On the left side, important dataset characteristics are described: sample size ($N$), no. of fixed effects features ($D$), no. of classes ($C$), no. of random effects features ($L$), and no. of clusters of the feature with the highest cardinality ($Q_{max}$). The datasets are sorted by highest cardinality and displayed separated by binary and multi-class targets. On the right side, the performance (AUC) of the approaches is compared. Results for the best method and results that do not significantly differ in a paired t-test ($\alpha=0.05$) are highlighted.}
    \label{tab:pargent}
\end{table*}

\paragraph{Experimental Setting} 
To evaluate the performance of MC-GMENN on real-world classification tasks, we use the same datasets as in \citeauthor{pargent2022regularized} [\citeyear{pargent2022regularized}], which is, to our knowledge, the largest benchmark study on the treatment of high-cardinality categorical features.
We use the same preprocessing as \citeauthor{pargent2022regularized} [\citeyear{pargent2022regularized}], where $Q=10$ is used as a threshold to define high-cardinality features. Low-cardinality categorical features are encoded using OHE. Other categorical features are treated as clustering features. 
More details on the datasets and preprocessing can be found in the supplementary material and in the paper of \citeauthor{pargent2022regularized} [\citeyear{pargent2022regularized}].
For all approaches, we use the neural network architecture implemented in the AutoML framework AutoGluon-tabular with its default hyperparameters \cite{erickson2020autogluon}. 

\begin{figure}[t!]
  \centering
  \small
  \includegraphics[width=0.9\columnwidth]{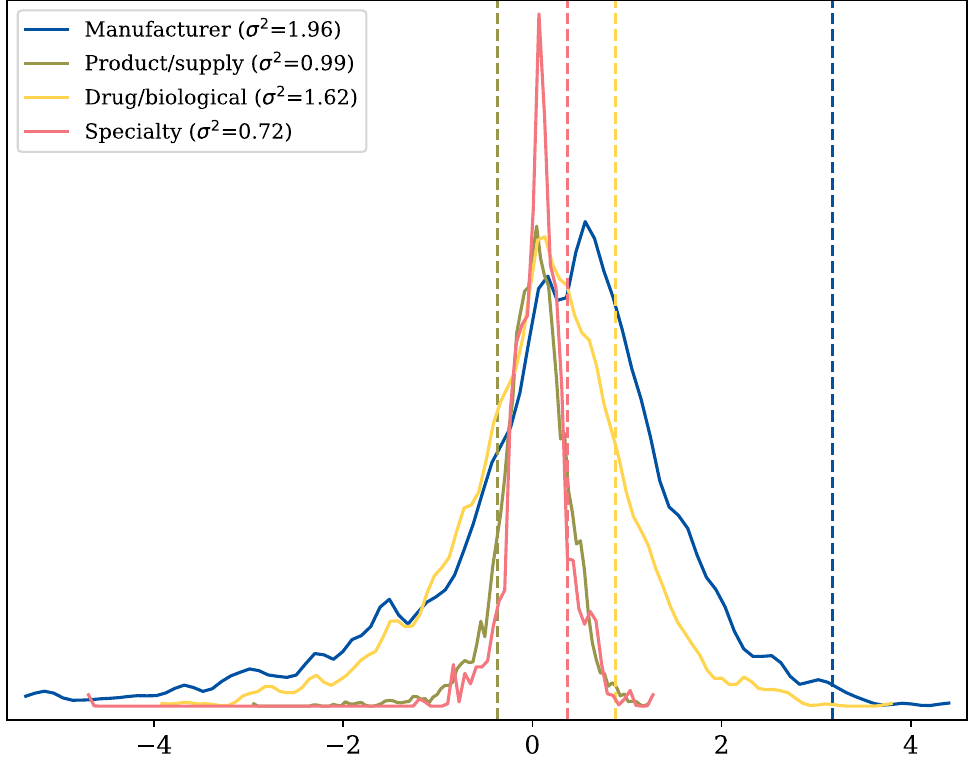}
  \caption{Interpretation example for learned random effects distributions on the \textit{open\_payments} dataset. The task is to classify whether payments from manufacturers to physicians are disallowed. The payments are clustered by manufacturer (Q=1460), the device associated with the payment (Q=4365), the drug associated with the payment (Q=2255), and the physician's specialty (Q=513). The dashed lines show estimated random effects for one example payment.}
   \label{fig:interpret_example}
\end{figure}

\paragraph{MC-GMENN Consistently Matches or Outperforms Encoding and Embedding Approaches} 
The results in Table \ref{tab:pargent} demonstrate strong performance of MC-GMENN across various real-world datasets. As in the previous subsection, MC-GMENN particularly excels on datasets with numerous high-cardinality clustering features and classes. Even for datasets where MC-GMENN does not achieve the top rank, the performance gap to the best model remains small. 
For some datasets, the \textit{Ignore} method ranks among the top-performing models. The fact that MC-GMENN exhibits comparable performance on most of these datasets highlights its strong capability in discerning irrelevant clustering features, which is an important property for tabular deep learning \cite{grinsztajn2022tree}.
In conclusion, MC-GMENN more consistently achieves high performance than competitive approaches.

\paragraph{MC-GMENN Enables Interpretability of Cluster Effects}
Figure \ref{fig:interpret_example} illustrates the interpretability of MC-GMENN on an example. The estimated random effects distribution can be used to assess global clustering feature importance. 
The estimated variances indicate that all four categorical features have clusters with impact, while manufacturer and drug have generally the most widespread distributions and thus potentially stronger effects.
As the random effects are linear, they provide white box access to the model behavior w.r.t. to the categorical features and can be used for local interpretability. 
For the illustrated sample, the payment is classified as disallowed. 
The model has estimated a high random effect on the logits for the manufacturer making the payment. Hence, the classification is greatly influenced by the fact that the payment is made by this particular manufacturer.

\section{Related Work}\label{sec:related-work}
In this section, we discuss related work with a focus on mixed effects deep learning approaches for classification. We refer to \cite{prokhorenkova2018catboost,pargent2022regularized} for the treatment of categorical data in general and to \cite{hancock2020survey,borisov2021deep,huang2020tabtransformer} for examples of how categorical data is typically treated in deep learning.
Existing MENN approaches are different from each other mainly in terms of the training procedure and their applicability, as summarized in Table \ref{tab:me-theoretical-comp}.

\citeauthor{xiong2019mixed} [\citeyear{xiong2019mixed}] propose \textbf{MeNets}, a MENN relying on variational EM with stochastic gradient descent that was originally proposed for regression. 
Notably, our EM framework is fundamentally different, as \citeauthor{xiong2019mixed} [\citeyear{xiong2019mixed}] update the fixed effects parameters in the E-step and rely on the expensive inversion of large covariance matrices to update the random effects. In contrast, we use MCMC to obtain random effect samples in the E-step and perform all parameter updates in the M-step. 
\citeauthor{nguyen2023adversarially} [\citeyear{nguyen2023adversarially}] showed that for binary classification, the approach is less efficient and less performant than LMMNN and ARMED, therefore we did not include it in our evaluation. 

\citeauthor{tran2020bayesian} [\citeyear{tran2020bayesian}] propose \textbf{DeepGLMM}, a MENN for panel data trained with Bayesian variational approximation.
Although theoretically widely applicable, the approach was only evaluated in a limited setting on one real-world dataset without comparison to other deep learning methods. Furthermore, the authors of recent MENN approaches were not able to use the implementation due its computational and conceptual complexity \cite{simchoni2023integrating,nguyen2023adversarially}. In contrast, we demonstrated that MC-GMENN is easily applicable to diverse datasets even with the default hyperparameters.

\citeauthor{simchoni2021using} [\citeyear{simchoni2021using}] focus on regression and develop a 
plug-in loss function to train linear mixed model neural networks (\textbf{LMMNN}). To this end, an important aspect is the decomposability of the loss function into batches. Remarkably, in our MCEM procedure, the fixed effects parameters can naturally be updated in mini-batches as the variance components usually hindering the decomposability are updated separately and conventional loss functions can be used. 
Recently, \citeauthor{simchoni2023integrating} [\citeyear{simchoni2023integrating}] extended their framework to binary classification with one clustering feature by leveraging Gaussian quadrature to integrate over the random effects. 
However, unlike our approach, LMMNN is not applicable to classification with multiple clustering features and classes due to the inability of Gaussian quadrature to scale effectively to high-dimensional integrals \cite{bolker2009generalized}. 
\textbf{ARMED} \cite{nguyen2023adversarially} 
is distinguished by the use of two additional networks besides the fixed effects network: one for predicting the effects of unseen clusters and one adversarial network to better disentangle clustering effects from the fixed effects.
The model is trained using VI 
and evaluated solely for binary classification with a single random effect and cardinalities up to 20. 
Using high-cardinality features as random effects leads to a parameter explosion of the two additional networks, greatly increasing training time and memory requirements. 
Furthermore, the bias towards the prior distribution prevents estimating larger random effects, which leads to poor performance for high-variance scenarios.
Our experiments in Subsection \ref{ssec:comp_me} have highlighted these limitations. Additionally, they have demonstrated that our approach scales effectively to high-dimensional random effects while offering accurate inter-cluster variance quantification.

Other existing approaches are limited to very small neural network architectures \cite{tandon2006neural,mandel2021neural} or have been only implemented for regression tasks \cite{simchoni2021using,avanzi2023machine,kilian2023mixed}. 
In addition, there are deep mixed effects models which use point estimates of random effects and thus are not comparable to the MENNs in our scope \cite{shi2022generalized,wortwein2023neural}.
Furthermore, various approaches combining GLMMs and tree-based models were proposed \cite{sela2012re,hajjem2017generalized,ngufor2019mixed,sigrist2022}.



As highlighted in Table \ref{tab:me-theoretical-comp}, all existing approaches come with restrictions in applicability. Our approach is the first that applies to multi-class classification datasets with multiple random effects. 
It is worth noting that some of the existing approaches may in theory be extendable. However, none of the available implementations is readily applicable to the datasets we considered in Subsections \ref{ssec:multi} and \ref{ssec:highcard} and the required modifications are nontrivial. 
Despite the discussed limitations, we want to emphasize that all approaches have particular strengths for specific scenarios. I.e., DeepGLMM can be applied to panel data, LMMNN is remarkably well suited for regression tasks and ARMED is the most efficient for low-dimensional random effects.
The main limitation of all mixed-effects approaches is that they introduce components that increase the training time compared to conventional encoding or embedding approaches. MeNets and LMMNN rely on expensive matrix inversions, ARMED introduces two additional networks, and DeepGLMM as well as MC-GMENN rely on sampling the posterior. 
Nevertheless, we have shown that our approach is efficient compared to others in high-cardinality scenarios and has the widest range of applicability.

\begin{table}[tb]
    \centering
    \small
    \begin{tabular}{rccccc}
    \toprule
    Approach & Training &   $C>2$ &     high $Q$ &      $L>1$ \\
    \midrule
    LMMNN & GQ & \color{red}\ding{55}  &    \color{green}\checkmark  &    {\color{red}\ding{55}}$^{1}$\\
    ARMED & VI & {\color{red}\ding{55}}  &    {\color{red}\ding{55}}  &    \color{red}\ding{55}   \\
    MeNets & VI & \color{red}\ding{55}   &   {\color{red}\ding{55}}     &     \color{red}\ding{55}  \\
    DeepGLMM & VI & \color{red}\ding{55}   &  \color{green}\checkmark   &   \color{red}\ding{55}    \\
    MC-GMENN (ours) & MCMC & \color{green}\checkmark    &  \color{green}\checkmark    &   \color{green}\checkmark    \\
    
    \bottomrule
    \end{tabular}
    \caption{Comparison of different MENN approaches for classification w.r.t. their training method and demonstrated application to multi-class targets ($C>2$), high-cardinality categorical features (high $Q$), and multiple random effects ($L>1$). $^{1}$only for regression.}
    \label{tab:me-theoretical-comp}
\end{table}

\section{Conclusion} \label{sec:conclusion}
In this paper, we proposed MC-GMENN, a novel framework for training generalized mixed effects neural networks using Monte Carlo methods. 
We have shown that due to the partially Bayesian nature of MENNs, Monte Carlo methods can be utilized with competitive time efficiency.
By decoupling batch updates from sampling in an MCEM procedure and state-of-the-art MCMC sampling techniques, random intercept neural networks for high-cardinality clustering features can be trained even more efficiently than using previous approaches. 
Furthermore, we demonstrated that our approach is able to correctly quantify inter-cluster variance, while previous approaches are biased and often unable to estimate the true posterior.
Our contribution allows to apply mixed effects neural networks to a wide range of classification problems, including multiple classes and clustering features. 
Future work includes investigating whether MC-GMENN can improve over state-of-the-art approaches in specific domains, such as medicine, click-through rate prediction, or human-centered data applications. 
Moreover, we hope that our work inspires researchers to challenge assumptions about the scalability of Monte Carlo methods for deep learning applications in other fields than mixed effects modeling.

\FloatBarrier





\bibliographystyle{named}
\bibliography{ijcai24}

\begin{thebibliography}{}

\bibitem[\protect\citeauthoryear{Agresti}{2012}]{agresti2012categorical}
Alan Agresti.
\newblock {\em Categorical data analysis}, volume 792.
\newblock John Wiley \& Sons, 2012.

\bibitem[\protect\citeauthoryear{Archila}{2016}]{archila2016markov}
Felipe Humberto~Acosta Archila.
\newblock {\em Markov chain Monte Carlo for linear mixed models}.
\newblock PhD thesis, university of minnesota, 2016.

\bibitem[\protect\citeauthoryear{Avanzi \bgroup \em et al.\egroup }{2023}]{avanzi2023machine}
Benjamin Avanzi, Greg Taylor, Melantha Wang, and Bernard Wong.
\newblock Machine learning with high-cardinality categorical features in actuarial applications.
\newblock {\em arXiv preprint arXiv:2301.12710}, 2023.

\bibitem[\protect\citeauthoryear{Blei \bgroup \em et al.\egroup }{2017}]{blei2017variational}
David~M Blei, Alp Kucukelbir, and Jon~D McAuliffe.
\newblock Variational inference: A review for statisticians.
\newblock {\em Journal of the American statistical Association}, 112(518):859--877, 2017.

\bibitem[\protect\citeauthoryear{Bolker \bgroup \em et al.\egroup }{2009}]{bolker2009generalized}
Benjamin~M Bolker, Mollie~E Brooks, Connie~J Clark, Shane~W Geange, John~R Poulsen, M~Henry~H Stevens, and Jada-Simone~S White.
\newblock Generalized linear mixed models: a practical guide for ecology and evolution.
\newblock {\em Trends in ecology \& evolution}, 24(3):127--135, 2009.

\bibitem[\protect\citeauthoryear{Borisov \bgroup \em et al.\egroup }{2021}]{borisov2021deep}
Vadim Borisov, Tobias Leemann, Kathrin Se{\ss}ler, Johannes Haug, Martin Pawelczyk, and Gjergji Kasneci.
\newblock Deep neural networks and tabular data: A survey.
\newblock {\em arXiv preprint arXiv:2110.01889}, 2021.

\bibitem[\protect\citeauthoryear{Cafri \bgroup \em et al.\egroup }{2019}]{cafri2019review}
Guy Cafri, Wei Wang, Priscilla~H Chan, and Peter~C Austin.
\newblock A review and empirical comparison of causal inference methods for clustered observational data with application to the evaluation of the effectiveness of medical devices.
\newblock {\em Statistical Methods in Medical Research}, 28(10-11):3142--3162, 2019.

\bibitem[\protect\citeauthoryear{Erickson \bgroup \em et al.\egroup }{2020}]{erickson2020autogluon}
Nick Erickson, Jonas Mueller, Alexander Shirkov, Hang Zhang, Pedro Larroy, Mu~Li, and Alexander Smola.
\newblock Autogluon-tabular: Robust and accurate automl for structured data.
\newblock {\em arXiv preprint arXiv:2003.06505}, 2020.

\bibitem[\protect\citeauthoryear{Fei \bgroup \em et al.\egroup }{2021}]{fei2021promoting}
Mengqi Fei, Huizhong Tan, Xixian Peng, Qiuzhen Wang, and Lei Wang.
\newblock Promoting or attenuating? an eye-tracking study on the role of social cues in e-commerce livestreaming.
\newblock {\em Decision Support Systems}, 142:113466, 2021.

\bibitem[\protect\citeauthoryear{Grinsztajn \bgroup \em et al.\egroup }{2022}]{grinsztajn2022tree}
L{\'e}o Grinsztajn, Edouard Oyallon, and Ga{\"e}l Varoquaux.
\newblock Why do tree-based models still outperform deep learning on typical tabular data?
\newblock {\em Advances in Neural Information Processing Systems}, 35:507--520, 2022.

\bibitem[\protect\citeauthoryear{Guo and Berkhahn}{2016}]{guo2016entity}
Cheng Guo and Felix Berkhahn.
\newblock Entity embeddings of categorical variables.
\newblock {\em arXiv preprint arXiv:1604.06737}, 2016.

\bibitem[\protect\citeauthoryear{Hajjem \bgroup \em et al.\egroup }{2017}]{hajjem2017generalized}
Ahlem Hajjem, Denis Larocque, and Fran{\c{c}}ois Bellavance.
\newblock Generalized mixed effects regression trees.
\newblock {\em Statistics \& Probability Letters}, 126:114--118, 2017.

\bibitem[\protect\citeauthoryear{Hancock and Khoshgoftaar}{2020}]{hancock2020survey}
John~T Hancock and Taghi~M Khoshgoftaar.
\newblock Survey on categorical data for neural networks.
\newblock {\em Journal of Big Data}, 7(1):1--41, 2020.

\bibitem[\protect\citeauthoryear{Harrison \bgroup \em et al.\egroup }{2018}]{harrison2018brief}
Xavier~A Harrison, Lynda Donaldson, Maria~Eugenia Correa-Cano, Julian Evans, David~N Fisher, Cecily~ED Goodwin, Beth~S Robinson, David~J Hodgson, and Richard Inger.
\newblock A brief introduction to mixed effects modelling and multi-model inference in ecology.
\newblock {\em PeerJ}, 6:e4794, 2018.

\bibitem[\protect\citeauthoryear{Hoffman and Gelman}{2011}]{hoffman2011nuts}
Matthew~D. Hoffman and Andrew Gelman.
\newblock The no-u-turn sampler: Adaptively setting path lengths in hamiltonian monte carlo, 2011.

\bibitem[\protect\citeauthoryear{Huang \bgroup \em et al.\egroup }{2020}]{huang2020tabtransformer}
Xin Huang, Ashish Khetan, Milan Cvitkovic, and Zohar Karnin.
\newblock Tabtransformer: Tabular data modeling using contextual embeddings.
\newblock {\em arXiv preprint arXiv:2012.06678}, 2020.

\bibitem[\protect\citeauthoryear{Jospin \bgroup \em et al.\egroup }{2022}]{jospin2022hands}
Laurent~Valentin Jospin, Hamid Laga, Farid Boussaid, Wray Buntine, and Mohammed Bennamoun.
\newblock Hands-on bayesian neural networks—a tutorial for deep learning users.
\newblock {\em IEEE Computational Intelligence Magazine}, 17(2):29--48, 2022.

\bibitem[\protect\citeauthoryear{Kilian \bgroup \em et al.\egroup }{2023}]{kilian2023mixed}
Pascal Kilian, Sangbeak Ye, and Augustin Kelava.
\newblock Mixed effects in machine learning--a flexible mixedml framework to add random effects to supervised machine learning regression.
\newblock {\em Transactions on Machine Learning Research}, 2023.

\bibitem[\protect\citeauthoryear{Mandel \bgroup \em et al.\egroup }{2021}]{mandel2021neural}
Francesca Mandel, Riddhi~Pratim Ghosh, and Ian Barnett.
\newblock Neural networks for clustered and longitudinal data using mixed effects models.
\newblock {\em Biometrics}, 2021.

\bibitem[\protect\citeauthoryear{McCulloch}{1997}]{mcculloch1997maximum}
Charles~E McCulloch.
\newblock Maximum likelihood algorithms for generalized linear mixed models.
\newblock {\em Journal of the American statistical Association}, 92(437):162--170, 1997.

\bibitem[\protect\citeauthoryear{McCulloch}{2003}]{mcculloch2003generalized}
Charles~E McCulloch.
\newblock Generalized linear mixed models.
\newblock Ims, 2003.

\bibitem[\protect\citeauthoryear{Micci-Barreca}{2001}]{micci2001preprocessing}
Daniele Micci-Barreca.
\newblock A preprocessing scheme for high-cardinality categorical attributes in classification and prediction problems.
\newblock {\em ACM SIGKDD Explorations Newsletter}, 3(1):27--32, 2001.

\bibitem[\protect\citeauthoryear{Monnahan and Kristensen}{2018}]{monnahan2018no}
Cole~C Monnahan and Kasper Kristensen.
\newblock No-u-turn sampling for fast bayesian inference in admb and tmb: Introducing the adnuts and tmbstan r packages.
\newblock {\em PloS one}, 13(5):e0197954, 2018.

\bibitem[\protect\citeauthoryear{Ngufor \bgroup \em et al.\egroup }{2019}]{ngufor2019mixed}
Che Ngufor, Holly Van~Houten, Brian~S Caffo, Nilay~D Shah, and Rozalina~G McCoy.
\newblock Mixed effect machine learning: a framework for predicting longitudinal change in hemoglobin a1c.
\newblock {\em Journal of biomedical informatics}, 89:56--67, 2019.

\bibitem[\protect\citeauthoryear{Nguyen \bgroup \em et al.\egroup }{2023}]{nguyen2023adversarially}
Kevin~P Nguyen, Alex~H Treacher, and Albert~A Montillo.
\newblock Adversarially-regularized mixed effects deep learning (armed) models improve interpretability, performance, and generalization on clustered (non-iid) data.
\newblock {\em IEEE Transactions on Pattern Analysis and Machine Intelligence}, 2023.

\bibitem[\protect\citeauthoryear{Pargent \bgroup \em et al.\egroup }{2022}]{pargent2022regularized}
Florian Pargent, Florian Pfisterer, Janek Thomas, and Bernd Bischl.
\newblock Regularized target encoding outperforms traditional methods in supervised machine learning with high cardinality features.
\newblock {\em Computational Statistics}, pages 1--22, 2022.

\bibitem[\protect\citeauthoryear{Pinheiro and Chao}{2006}]{pinheiro2006efficient}
Jos{\'e}~C Pinheiro and Edward~C Chao.
\newblock Efficient laplacian and adaptive gaussian quadrature algorithms for multilevel generalized linear mixed models.
\newblock {\em Journal of Computational and Graphical Statistics}, 15(1):58--81, 2006.

\bibitem[\protect\citeauthoryear{Prokhorenkova \bgroup \em et al.\egroup }{2018}]{prokhorenkova2018catboost}
Liudmila Prokhorenkova, Gleb Gusev, Aleksandr Vorobev, Anna~Veronika Dorogush, and Andrey Gulin.
\newblock Catboost: unbiased boosting with categorical features.
\newblock {\em Advances in neural information processing systems}, 31, 2018.

\bibitem[\protect\citeauthoryear{Sela and Simonoff}{2012}]{sela2012re}
Rebecca~J Sela and Jeffrey~S Simonoff.
\newblock Re-em trees: a data mining approach for longitudinal and clustered data.
\newblock {\em Machine learning}, 86(2):169--207, 2012.

\bibitem[\protect\citeauthoryear{Shi \bgroup \em et al.\egroup }{2022}]{shi2022generalized}
Jun Shi, Chengming Jiang, Aman Gupta, Mingzhou Zhou, Yunbo Ouyang, Qiang~Charles Xiao, Qingquan Song, Yi~Wu, Haichao Wei, and Huiji Gao.
\newblock Generalized deep mixed models.
\newblock In {\em Proceedings of the 28th ACM SIGKDD Conference on Knowledge Discovery and Data Mining}, pages 3869--3877, 2022.

\bibitem[\protect\citeauthoryear{Sigrist}{2022}]{sigrist2022}
Fabio Sigrist.
\newblock Gaussian process boosting.
\newblock {\em Journal of Machine Learning Research}, 23(232):1--46, 2022.

\bibitem[\protect\citeauthoryear{Sigrist}{2023}]{sigrist2023comparison}
Fabio Sigrist.
\newblock A comparison of machine learning methods for data with high-cardinality categorical variables.
\newblock {\em arXiv preprint arXiv:2307.02071}, 2023.

\bibitem[\protect\citeauthoryear{Simchoni and Rosset}{2021}]{simchoni2021using}
Giora Simchoni and Saharon Rosset.
\newblock Using random effects to account for high-cardinality categorical features and repeated measures in deep neural networks.
\newblock {\em Advances in Neural Information Processing Systems}, 34:25111--25122, 2021.

\bibitem[\protect\citeauthoryear{Simchoni and Rosset}{2023}]{simchoni2023integrating}
Giora Simchoni and Saharon Rosset.
\newblock Integrating random effects in deep neural networks.
\newblock {\em Journal of Machine Learning Research}, 24(156):1--57, 2023.

\bibitem[\protect\citeauthoryear{Tandon \bgroup \em et al.\egroup }{2006}]{tandon2006neural}
Reeti Tandon, Sudeshna Adak, and Jeffrey~A Kaye.
\newblock Neural networks for longitudinal studies in alzheimer’s disease.
\newblock {\em Artificial intelligence in medicine}, 36(3):245--255, 2006.

\bibitem[\protect\citeauthoryear{Tran \bgroup \em et al.\egroup }{2020}]{tran2020bayesian}
M-N Tran, Nghia Nguyen, David Nott, and Robert Kohn.
\newblock Bayesian deep net glm and glmm.
\newblock {\em Journal of Computational and Graphical Statistics}, 29(1):97--113, 2020.

\bibitem[\protect\citeauthoryear{W{\"o}rtwein \bgroup \em et al.\egroup }{2023}]{wortwein2023neural}
Torsten W{\"o}rtwein, Nicholas~B Allen, Lisa~B Sheeber, Randy~P Auerbach, Jeffrey~F Cohn, and Louis-Philippe Morency.
\newblock Neural mixed effects for nonlinear personalized predictions.
\newblock In {\em Proceedings of the 25th International Conference on Multimodal Interaction}, pages 445--454, 2023.

\bibitem[\protect\citeauthoryear{Xiong \bgroup \em et al.\egroup }{2019}]{xiong2019mixed}
Yunyang Xiong, Hyunwoo~J Kim, and Vikas Singh.
\newblock Mixed effects neural networks (menets) with applications to gaze estimation.
\newblock In {\em Proceedings of the IEEE/CVF conference on computer vision and pattern recognition}, pages 7743--7752, 2019.

\end{thebibliography}

\end{document}